\documentclass[10pt,twocolumn,letterpaper,table]{article}

\usepackage{wacv}      

\usepackage{amsmath,amssymb,amsfonts}
\usepackage{algorithmic}
\usepackage{graphicx}
\usepackage{textcomp}

\usepackage{algorithm}
\usepackage{array}
\usepackage{textcomp}
\usepackage{stfloats}
\usepackage{url}
\usepackage{verbatim}
\usepackage{subcaption}

\usepackage{booktabs}
\usepackage[accsupp]{axessibility}  
\usepackage{siunitx}

\usepackage{tikz}
\definecolor{myblue}{HTML}{3333FF}
\definecolor{mygreen}{HTML}{D5E8D4}
\definecolor{mysalmon}{HTML}{F8CECC}
\definecolor{mybrown}{HTML}{FAD7AC}
\definecolor{myblueish}{HTML}{DAE8FC}
\definecolor{mygray}{HTML}{BAC8D3}
\definecolor{mypurple}{HTML}{D0CEE2}
\definecolor{myyellow}{HTML}{FFF2CC}

\newcommand{\specialcell}[2][l]{%
  \begin{tabular}[#1]{@{}l@{}}#2\end{tabular}}

\newcommand{\sphere}[1]{
    \begin{tikzpicture}[scale=#1]
        \shade[ball color = myyellow, opacity = 1] (0,0) circle (1cm);
        \draw (0,0) circle (1cm);
        \draw (-1,0) arc (180:360:1 and 0.3);
        \draw[dashed] (1,0) arc (0:180:1 and 0.3);
    \end{tikzpicture}
}

\newcommand{\cube}[1]{
    \begin{tikzpicture}[scale=#1]
        \draw[fill=myyellow] (0,0) rectangle (1,1); 
        \draw[fill=myyellow] (0.3,0.3) rectangle (1.3,1.3); 
        \draw[fill=myyellow] (0,0) -- (0.3,0.3) -- (1.3,0.3) -- (1,0) -- cycle; 
        \draw[fill=myyellow] (1,1) -- (1.3,1.3) -- (1.3,0.3) -- (1,0) -- cycle; 
        \draw[fill=myyellow] (0,1) -- (0.3,1.3) -- (1.3,1.3) -- (1,1) -- cycle; 
    \end{tikzpicture}
}

\newcommand{\Oshape}[1]{
    \begin{tikzpicture}[scale=#1]
        \draw (0,0) circle (1.5cm); 
        \foreach \a in {0,30,...,330} {
          \fill (\a:1.5cm) circle (10pt); 
        }
    \end{tikzpicture}
}

\newcommand{\OshapeSix}[1]{
    \begin{tikzpicture}[scale=#1]
        \draw (0,0) circle (1.5cm); 
        \foreach \a in {0,60,...,300} {
          \fill (\a:1.5cm) circle (10pt); 
        }
    \end{tikzpicture}
}

\newcommand{\Ushape}[1]{
    \begin{tikzpicture}[scale=#1]
        \draw (0,3) -- (0,1.5) -- (0,0) -- (1.5,0) -- (3,0) -- (3,1.5) -- (3,3); 
        \foreach \x in {0,1.5,...,3} {
          \fill (\x,0) circle (10pt);
        }
        \fill (0,3) circle (10pt);
        \fill (0,1.5) circle (10pt);
        \fill (3,1.5) circle (10pt);
        \fill (3,3) circle (10pt);
    \end{tikzpicture}
}

\newcommand{\Tshape}[1]{
    \begin{tikzpicture}[scale=#1]
        \draw (0,1.5) -- (1.5,1.5) -- (1.5,0) -- (1.5,1.5) -- (3,1.5);
        \foreach \x in {0,1.5,...,3} {
          \fill (\x,1.5) circle (10pt);
        }
        \fill (1.5,0) circle (10pt);
    \end{tikzpicture}
}

\usepackage[pagebackref,breaklinks,colorlinks]{hyperref}

\usepackage[capitalize]{cleveref}
\crefname{section}{Sec.}{Secs.}
\Crefname{section}{Section}{Sections}
\Crefname{table}{Table}{Tables}
\crefname{table}{Tab.}{Tabs.}

\def\wacvPaperID{1751} 
\def\confName{WACV}
\def\confYear{2025}

\begin{document}

\title{Neural Real-Time Recalibration for Infrared Multi-Camera Systems}

\author{Benyamin Mehmandar$\dagger$, Reza Talakoob$\dagger$, Charalambos Poullis\\
Immersive and Creative Technologies Lab\\
Concordia University\\
}
\maketitle
\renewcommand{\thefootnote}{\fnsymbol{footnote}}
\footnotetext[2]{These authors contributed equally to this work.}

\begin{abstract}
Currently, there are no learning-free or neural techniques for \textit{real-time} recalibration of \textit{infrared multi-camera} systems. In this paper, we address the challenge of real-time, highly-accurate calibration of multi-camera infrared systems, a critical task for time-sensitive applications. Unlike traditional calibration techniques that lack adaptability and struggle with on-the-fly recalibrations, we propose a neural network-based method capable of dynamic real-time calibration. The proposed method integrates a differentiable projection model that directly correlates 3D geometries with their 2D image projections and facilitates the direct optimization of both intrinsic and extrinsic camera parameters. Key to our approach is the dynamic camera pose synthesis with perturbations in camera parameters, emulating realistic operational challenges to enhance model robustness. We introduce two model variants: one designed for multi-camera systems with onboard processing of 2D points, utilizing the direct 2D projections of 3D fiducials, and another for image-based systems, employing color-coded projected points for implicitly establishing correspondence. 
Through rigorous experimentation, we demonstrate our method is more accurate than traditional calibration techniques \textit{with or without} perturbations while also being \textit{real-time}, marking a significant leap in the field of real-time multi-camera system calibration. The source code can be found at \url{https://github.com/theICTlab/neural-recalibration}
\end{abstract}

\section{Introduction}
\label{sec:intro}
Camera calibration is the fundamental process of estimating the camera's intrinsic and extrinsic parameters and is an essential part of many computer vision systems. Accurate calibration is important in various applications, ranging from 3D reconstruction to augmented reality, especially in settings demanding high accuracy, like surgical environments. Traditional calibration methods provide analytical frameworks for addressing camera calibration. However, they require capturing an object of known geometry from multiple viewpoints, then extracting points and establishing correspondences. The inherent computational complexity of these conventional methods tends to increase dramatically with the number of cameras, images, and correspondences, making them impractical for real-time applications.

In time-critical applications requiring high accuracy, standard commercial multi-camera systems have a fixed camera configuration and come pre-calibrated from manufacturers. However, the calibration of these systems deteriorates over time because of wear and tear and, more commonly, because of the buildup of debris on critical components like the fiducials or lenses. In the field of vision-based computer-assisted surgery, calibration problems frequently prevent multi-camera systems from meeting the exact specifications and stringent accuracy standards required for surgical procedures. Such discrepancies can compromise the effectiveness and safety of medical procedures. 

This deterioration in calibration emphasizes the need for sophisticated calibration methods. Coordinate Measuring Machines (CMM) are commonly used to improve calibration accuracy, offering a potential solution to this problem. Nonetheless, CMMs are expensive, and manufacturers typically calibrate the multi-camera system before delivery, failing to account for potential discrepancies after deployment. Traditional methods for detecting calibration errors, such as those based on epipolar geometry, face significant computational challenges in multi-camera setups and do not support on-the-fly recalibration, making them ineffective in dynamic environments.

\textbf{System Context and Problem Statement.} This work addresses the calibration challenges inherent in infrared multi-camera systems designed for time-critical applications, specifically vision-based computer-assisted surgery. These systems function exclusively within a fixed distance from a central point of interest, allowing rotation while maintaining a constant radius to ensure comprehensive visual coverage. This operational design is very important for three main reasons: first, to comply with the manufacturer's operating specifications and maintain system integrity and performance; second, to ensure that the area of interest is in-focus across all cameras for accurate tracking of medical instruments; and third, to minimize occlusions and reduce instances where tracked markers are not visible in all camera views. By customizing our approach to accommodate these multi-camera configurations and optimizing calibration for a fixed range of motion, we are able to significantly improve the accuracy and achieve the real-time calibration required for such time-critical applications.


In this paper, we introduce a novel real-time multi-camera calibration method that leverages neural networks to provide on-the-fly recalibration. 
Our model is trained on synthesized camera poses resulting from OEM calibration parameters, with perturbations applied to the intrinsic and extrinsic parameters. The perturbations emulate real-world operational challenges, thereby enhancing the model's practical applicability. We demonstrate, through rigorous experimentation, that our method not only adapts to alterations in calibration parameters in real time but also surpasses conventional calibration techniques in accuracy. Our key technical contributions are threefold: 
\begin{itemize}   
    \item First, we introduce a real-time neural calibration method for multi-camera systems, marking a departure from traditional offline calibration methods. Our method employs a differentiable projection model to flow gradients between 3D geometries and their 2D projections, allowing for direct optimization of camera parameters. 
    \item Second, we enhance the robustness and applicability of our method by dynamically synthesizing camera poses at each epoch and incorporating perturbations to the OEM calibration intrinsic and extrinsic parameters to simulate realistic operational challenges. 
    \item Finally, we introduce two variants of our model: the first is designed for multi-camera systems equipped with onboard processing, directly outputting the 2D projections of the 3D fiducials; the second variant is designed for image-based multi-camera systems.
\end{itemize}

The paper is structured as follows: Section \ref{sec:camera_model} provides a brief overview of the pinhole camera model. Section \ref{sec:methodology} discusses our methodology, including the dynamic camera pose synthesis for multi-camera systems and details on 
the loss functions and training strategy. Finally, Section \ref{sec:experiments} reports on our experimental results. 

\section{Related Work}

Camera calibration is fundamental in computer vision for determining geometric parameters essential for image capture, playing a vital role in applications that require accurate scene measurements. Despite the development of various methods to calculate camera parameters, traditional techniques like the Radial Distortion Model \cite{duane1971close}, Direct Linear Transform (DLT) \cite{abdeldirectlineartransformation}, Tsai's method \cite{tsai1987versatile}, Zhang's approach \cite{zhang2000flexible}, and \cite{shah1994simple} rely heavily on handcrafted features and model assumptions. These methods, while effective, are often labor-intensive and not suited for real-time or multi-camera calibration due to their complexity and the static nature of their required setup. The perspective-n-point (PnP) problem has seen advancements with \cite{nakano2016versatile}, \cite{urban2016mlpnp}, \cite{haner2015absolute}, and further improvements in \cite{cao2018fast} through direct optimization methods. However, these solutions still face challenges in terms of real-time execution.

The advent of artificial neural networks has prompted a significant shift in camera calibration research. Techniques like \cite{nguyen2022calibbd}, \cite{itu2017automatic}, and PoseNet \cite{kendall2015posenet} for extrinsic calibration and others \cite{workman2015deepfocal}, \cite{cramariuc2020learning}, \cite{hagemann2023deep}, \cite{bogdan2018deepcalib} for intrinsic parameter estimation leverage deep learning for more adaptable and potentially real-time calibration. Yet, these neural network approaches require extensive datasets for training, have low accuracy, and cannot generalize well across varying conditions.

In multi-camera calibration, research has extended towards self-calibration methods for setups in shared environments, as seen in the works of Svoboda \etal. \cite{svoboda2005convenient}, Heikkilä and Silvén \cite{heikkila1997four}, and also \cite{faugeras1992camera}, \cite{fraser1997digital}, \cite{hartley1994self}. These methods facilitate 3D reconstruction and multi-view analysis but remain challenged by the demands of real-time processing and dynamic scenes. Recent developments have also explored joint estimation of intrinsic and extrinsic parameters, as demonstrated by \cite{hold2018perceptual}, \cite{butt2022camera}, \cite{lopez2019deep}, \cite{wakai2021deep}.  These approaches promise more integrated calibration processes through deep learning, highlighting the potential for efficient real-time calibration. Nonetheless, achieving a balance between computational efficiency and accuracy remains a critical challenge for these advanced methods.



In contrast to the aforementioned techniques, our method enables real-time recalibration, effectively predicting camera parameters in the presence of perturbations to the OEM intrinsic calibration parameters. Additionally, a streamlined procedure for dynamic camera pose synthesis facilitates its generalization to arbitrary configurations of multi-camera systems and arbitrary calibration objects.

\section{Camera Model}
\label{sec:camera_model}
For the sake of completeness, this section provides an overview of the classic pinhole camera model, incorporating lens distortion to establish the relationship between the 3D world coordinates of a point $\mathbf{P}$ and its 2D image projection $\mathbf{p} = [x, y]$. Consider \( \mathbf{P} = [X, Y, Z]^T \) as a point in world coordinates, with \( \mathbf{R} \) and \( \mathbf{t} \) representing the camera's rotation matrix and translation vector, respectively. The transformation of the 3D point into camera coordinates \( \mathbf{P}^{C} = [X^{C}, Y^{C}, Z^{C}]^T\) is given by $\mathbf{P}^{C} = \mathbf{R} \mathbf{P} + \mathbf{t}$. The projection from 3D to 2D coordinates is given by:

\begin{equation}
  \begin{bmatrix}
    x \\
    y \\
    1 
  \end{bmatrix}
  =
  \frac{1}{Z^{C}}
  \begin{bmatrix}
    f_x & 0 & c_x \\
    0 & f_y & c_y \\
    0 & 0 & 1
  \end{bmatrix}
  \mathbf{P}^{C},
  \label{eq:projection}
\end{equation}
where \( Z^{C} \) is the third component of \( \mathbf{P}^{C} \), \( f_x \) and \( f_y \) the focal lengths, and \( c_x, c_y \) the principal point coordinates.

\noindent \textbf{Lens Distortion.} In order to accurately represent lens distortion, which commonly occurs in real camera systems, we include both the radial and tangential distortion. The distortion is represented using a polynomial model, where the radial distortion is captured by a sixth-order polynomial and the tangential distortion is addressed through first-order terms. The equations for the distorted coordinates $x_{\text{dist}}$ and $y_{\text{dist}}$ are given as follows:
\begin{equation}
  \begin{aligned}
    x_{\text{dist}} &= x (1 + k_1 r^2 + k_2 r^4 + k_3 r^6) + 2p_1xy + p_2(r^2 + 2x^2), \\
    y_{\text{dist}} &= y (1 + k_1 r^2 + k_2 r^4 + k_3 r^6) + p_1(r^2 + 2y^2) + 2p_2xy,
  \end{aligned}
\end{equation}
where \( r^2 = x^2 + y^2 \), and \( k_1, k_2, k_3 \) are the radial distortion coefficients, and \( p_1, p_2 \) are the tangential distortion coefficients.

\noindent \textbf{6D Rotation Parameterization.}
In line with the approach proposed in Zhou \etal. \cite{Zhou_2019_CVPR}, our work adopts a 6D parameterization for the representation of 3D rotations, diverging from traditional quaternion and Euler angle representations. The latter are known to introduce parameter space discontinuities, as depicted in Figure \ref{fig:discontinuity_diagram}, complicating the 

\noindent\begin{minipage}{0.5\textwidth}
learning process for neural networks due to the inherent discontinuities. Furthermore, a critical limitation of these traditional approaches is their inability to guarantee that the network outputs are orthogonal rotation matrices, which is often a desirable property for ensuring the physical plausibility of rotations in 3D space.  These limitations are particularly pronounced in tasks requiring the learning of
\end{minipage}%
\hfill
\begin{minipage}{0.45\textwidth}
    \captionsetup{skip=1pt}
    \centering 
    \includegraphics[width=\textwidth]{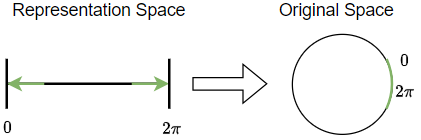}
    \captionof{figure}{Discontinuities introduced by quaternions and Euler angle representations hinder network learning efficiency as shown in \cite{Zhou_2019_CVPR}.}
    \label{fig:discontinuity_diagram}
\end{minipage}
\noindent continuous and smooth rotation spaces. For the purpose of training, a 3D rotation matrix $R$ is parameterized by a 6D vector $r_{6D}$, which encapsulates the elements of $R$'s first two columns. This parameterization provides a direct mapping to a rotation matrix without the need for complex conversions or the risk of introducing discontinuities. Specifically, during loss computation, the $r_{6D}$ vector is seamlessly transformed back into an orthogonal rotation matrix $R$. 
This approach not only aligns with the findings from \cite{Zhou_2019_CVPR} that direct regression on 3x3 rotation matrices can lead to larger errors, but also addresses the need for generating orthogonal matrices directly from the network.

\noindent \textbf{Differentiable Projection.}
The image formation process with the pinhole camera model is designed to be differentiable, facilitating the backpropagation of gradients from the loss -a function of the difference between the observed and projected points-  to the camera parameters. 

\section{Methodology}
\label{sec:methodology}
Our methodology employs a neural network designed to estimate the intrinsic and extrinsic parameters of a multi-camera system, leveraging a known calibration object in the scene. This tailored approach ensures that, once trained, the model is specifically attuned to the multi-camera system and calibration object utilized during its training phase. We invert the traditional image formation process, enabling our model to deduce camera parameters—such as rotations ($R$), translations ($t$), focal lengths ($f_x, f_y$), principal points ($c_x, c_y$), and distortion coefficients ($k_{1,2,3}, p_{1,2}$)—from 2D projections of 3D fiducial points on the calibration object.

\begin{figure*}[!ht]
    \centering
    \includegraphics[width=\textwidth]{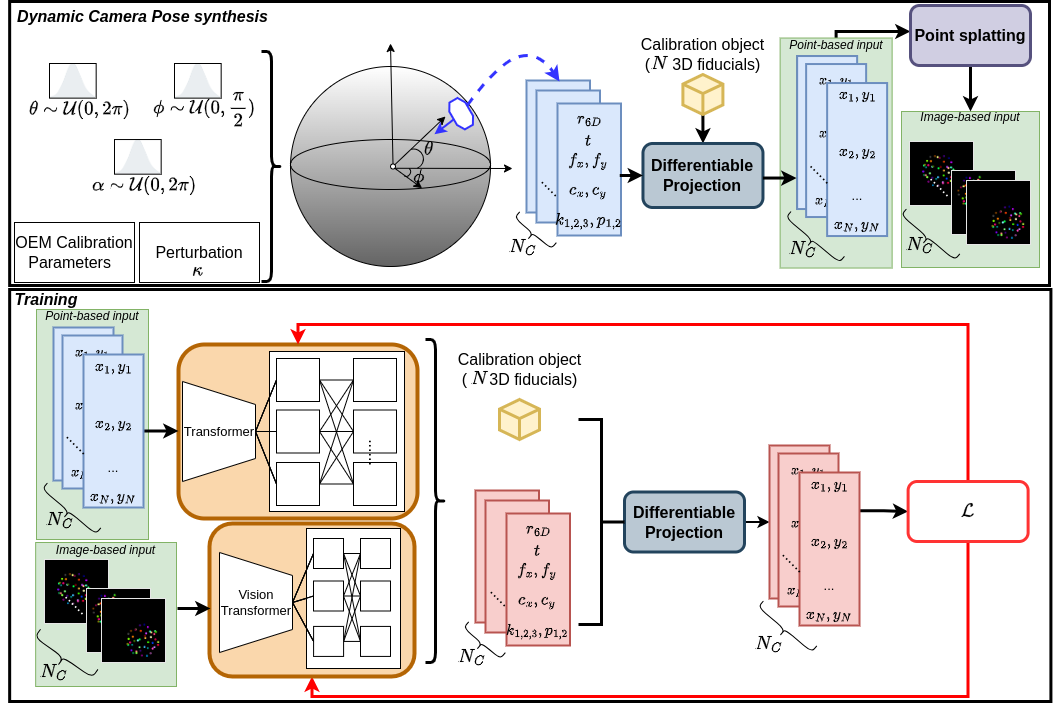}
    \caption{\textbf{Technical overview.} Our methodology begins with the synthesis of dynamic camera poses (see top fig.). Given spherical angles $\phi$ (azimuth), $\theta$ (elevation), along with the intrinsic rotation angle $\alpha$, the OEM calibration parameters, the maximum perturbation limit $\kappa$, and known 3D fiducials (e.g. a cube calibration object), this module performs two primary functions: (i) it {\setlength{\fboxsep}{0pt}\colorbox{myblueish}{synthesizes poses}} for the multi-camera system, and (ii) it computes the {\setlength{\fboxsep}{0pt}\colorbox{mygreen}{projected 2D points}}. Subsequently, it employs {\setlength{\fboxsep}{0pt}\colorbox{mypurple}{point splatting}} to render {\setlength{\fboxsep}{0pt}\colorbox{mygreen}{images}} of these points. During training (see bottom fig.), the {\setlength{\fboxsep}{0pt}\colorbox{myblueish}{synthesizes poses}} and {\setlength{\fboxsep}{0pt}\colorbox{mygreen}{projected points (alternatively rendered images)}} are used to train the {\setlength{\fboxsep}{0pt}\colorbox{mybrown}{neural network}}. A {\setlength{\fboxsep}{0pt}\colorbox{mygray}{differentiable projection}} ensures the propagation of gradients from the loss $\mathcal{L}$ back to the predicted camera parameters.}
    \label{fig:technical_overview}
\end{figure*}

At the core of our methodology is the dynamic camera pose synthesis, imperative for simulating realistic conditions that a multi-camera system might encounter. We introduce controlled perturbations into the OEM camera parameters, dynamically synthesizing diverse training samples. Given the synthesized camera parameters, the process starts with projecting the 3D fiducials of the calibration object onto the image plane. These 2D projected points serve as input to our model, which then predicts the intrinsics and extrinsics of the multi-camera system. Utilizing these predicted camera parameters, we project the known 3D fiducials back onto the image plane through a differentiable process. The deviation between the projections from the synthesized and predicted camera parameters is quantified using a loss function. 
Figure \ref{fig:technical_overview} provides an overview of our method.

\subsection{Dynamic Camera Pose Synthesis}
\label{subsec:train_data_generation}
Dynamic camera pose synthesis begins with the OEM calibration parameters of a multi-camera setup, typically determined by the manufacturing process. We represent a multi-camera setup of $N_{C}$ cameras as $C_{i}, 0 \leq i < N_{C}$, where $C_{i}^{OEM}$ denotes the initial calibration of each camera.

\subsubsection{Perturbations} To create a robust model capable of handling changes in calibration, we introduce perturbations to the OEM camera parameters, generating perturbed parameters $C_{i}^{pert} = C_{i}^{OEM} \times (1 + \delta)$, where $\delta$ represents the perturbation defined as $\delta = \kappa \times \mathcal{U}(0, 1)$, with $\kappa$ controlling the maximum desired perturbation. 

\subsubsection{Camera Pose Synthesis} For the synthesis of camera poses, we employ a two-step process, ensuring both focus towards the fiducial points and variability in camera orientation to prevent model overfitting. The procedure is as follows.

The centroid of the camera system is dynamically positioned on a hemisphere's surface, ensuring varied perspectives. The radius of the hemisphere is determined in advance based on the application requirements and the range of motion of the multi-camera system. Specifically, the centroid's position $P_{centroid}$ on the hemisphere is determined by:
\begin{equation}
P_{centroid} = (x_c, y_c, z_c) = \rho \cdot (\sin(\phi) \cdot \cos(\theta), \sin(\phi) \cdot \sin(\theta), \cos(\phi)),
\end{equation}
where $\rho$ denotes the hemisphere radius, and $\theta \sim \mathcal{U}(0, 2\pi), \phi \sim \mathcal{U}(0, \pi/2)$ are angles sampled from a uniform distribution, ensuring the centroid is randomly positioned over the hemisphere.

Next, a rotation $R_{focus}$ is applied to align the camera's viewing direction towards the fiducials' centroid, ensuring that the camera is oriented towards the area of interest. This alignment is critical for simulating realistic camera setups where the fiducials are within the camera's field of view. To introduce additional randomness and prevent the network from overfitting to specific camera locations, a secondary random rotation $R_{random}$ is applied by an intrinsic rotation angle $\alpha \sim \mathcal{U}(0, 2\pi)$ around the centroid point. The combination of $R_{focus}$ and $R_{random}$ ensures that each camera is not only oriented towards the fiducials but also positioned and rotated in a manner that provides a diverse set of viewing angles and positions. This diversity is imperative for training a robust model capable of generalizing across various camera orientations and positions.

As a final validation step, we conduct a visibility check to ensure all fiducials are within the field of view of all cameras. This step is essential, since the random placement of cameras on the hemisphere might result in scenarios where not all fiducials are visible from all cameras. 

\begin{figure*}[t]
\centering
\begin{tabular}{cccc}
\includegraphics[width=0.24\textwidth]{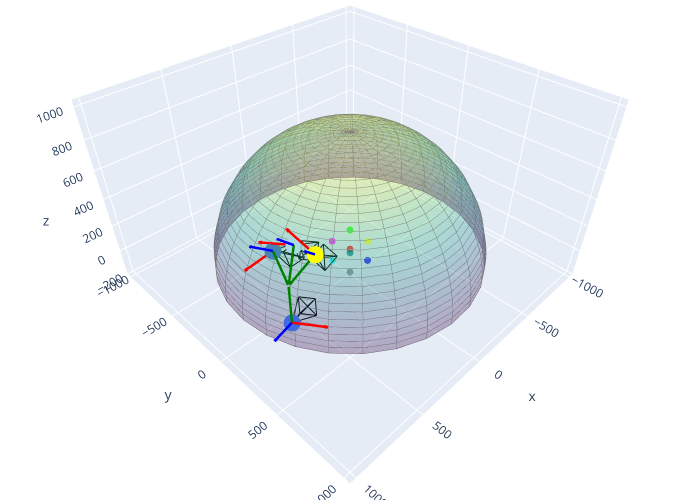} &
\includegraphics[width=0.24\textwidth]{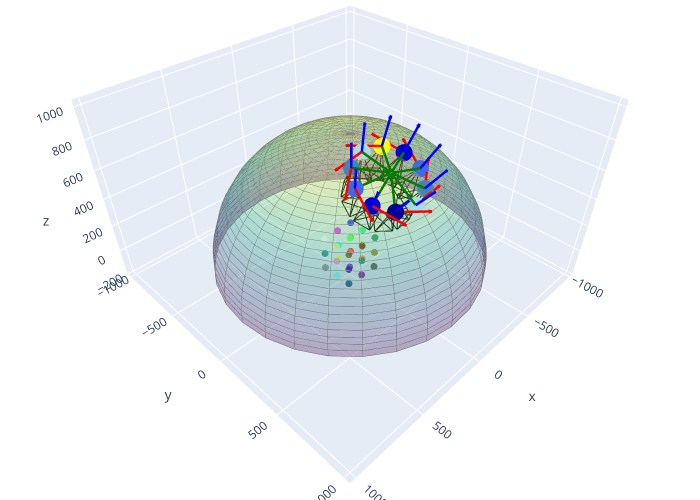} &
\includegraphics[width=0.24\textwidth]{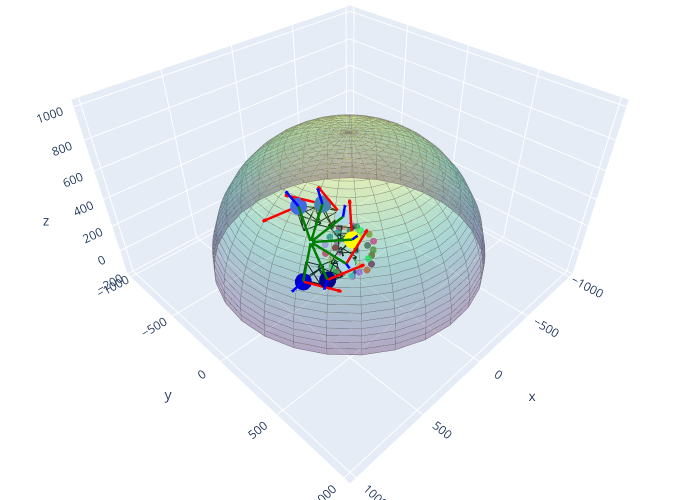} &
\includegraphics[width=0.24\textwidth]{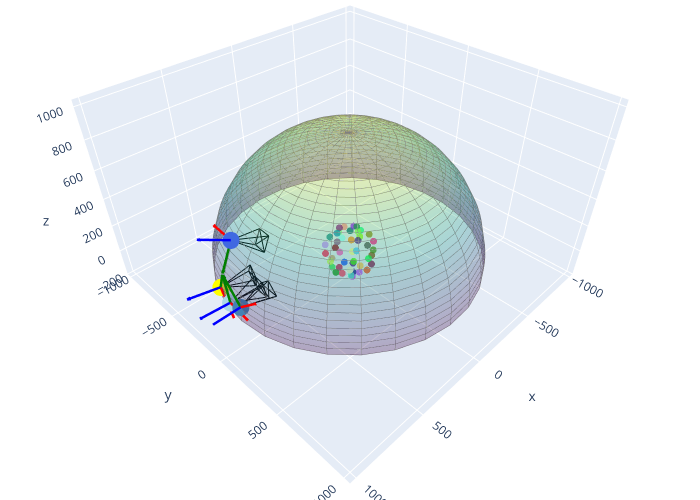} \\
(a) T; cube(8) & (b) O; cube(27) & (c) U; sphere(64) & (d) T; sphere(64) \\
\end{tabular}
\caption{\textbf{Dynamic Camera Pose Synthesis.} Our framework supports arbitrary configurations of multiple cameras as well as a wide range of calibration objects. To synthesize camera poses, we employ a random uniform sampling strategy across three dimensions to ensure a comprehensive exploration of the pose space: azimuth ($\theta$), elevation ($\phi$), and roll ($\alpha$), where $\theta \sim \mathcal{U}(0, 2\pi)$, $\phi \sim \mathcal{U}(0, \frac{\pi}{2})$, and $\alpha \sim \mathcal{U}(0, 2\pi)$. Additionally, Original Equipment Manufacturer (OEM) calibration parameters and a predefined maximum perturbation limit ($\kappa$) are incorporated.}
\label{fig:pose_generation}
\end{figure*}

The dynamic camera pose synthesis for each training epoch diversifies the dataset, improving model generalization. Figure \ref{fig:pose_generation} illustrates synthesized camera poses for various multi-camera systems e.g. T-shape ($N_{C}=4$ \Tshape{0.08}), U-shape ($N_{C}=7$ \Ushape{0.08}), O-shape ($N_{C}=10$ \Oshape{0.08}), and different calibration objects e.g. cube ($N_{fid.} = 8$), cube ($N_{fid.} = 27$), sphere ($N_{fid.} = 64$).

\subsection{Network Architecture}
\label{sec:model}
We introduce two variants: the first one, known as the point-based model (Pt), is intended for multi-camera systems with onboard processing that directly outputs the 2D projections of the 3D fiducials, which is the typical case for IR multi-camera systems; the second one is the image-based model (Img), designed for multi-camera systems that output images for subsequent processing, rather than the point coordinates. The architecture of both variants is detailed in the supplementary material along with an ablation study on the components.

\subsection{Loss}
Our loss function is crafted to encapsulate multiple facets of camera parameter estimation. The primary term $\mathcal{L}{\epsilon}$ is incorporated to enforce geometric consistency. It measures the RMSE between 2D points $x, x'$ obtained by projecting the 3D points using the predicted and ground-truth parameters, respectively, and is given by $\mathcal{L}{\epsilon} = RMSE(x, x')$. For rotation, we use geodesic loss $\mathcal{L}_{geo}$, computed from the predicted and ground-truth rotation matrices, to penalize deviations in the orientation. 
Additionally, $\mathcal{L}_{diff}$ is the root mean square error (RMSE) that quantifies the difference between the predicted and ground-truth parameters. We also employ a scaling value for the rotation and distortion coefficients since their values are very small compared to other camera parameters, as illustrated by the experiments in the ablations in the supplementary material.

\subsection{Training}
\label{sec:training}
In the initial phase of 10,000 epochs, we utilize a simplified loss function focused on $\mathcal{L}_{diff}$ and $\mathcal{L}_{geo}$ to establish a robust baseline, formulated as $ \mathcal{L} = \lambda_1 \times \mathcal{L}_{diff} + \lambda_1 \times \mathcal{L}_{geo}$. Subsequently, we introduce the full compound loss, expressed as $ \mathcal{L} = \lambda_1 \times \mathcal{L}_{diff} + \lambda_1 \times \mathcal{L}_{geo} + \lambda_2 \times \mathcal{L}_{\epsilon}$. In all experiments, the coefficients are set to $\lambda_1 = 100, \lambda_2 = 0.01$, with the batch size of 512 and 8 for Pt and Img variants, respectively.


Our training strategy is based on introducing a random perturbation to the camera parameters in every epoch, to simulate a wide spectrum of deviations that may occur due to buildup of debris on critical components like the fiducials or lenses. This methodical addition of perturbation aids in better exploration of the parameter space and enhances the robustness and accuracy of the estimated camera parameters. We employ an Adam optimizer with parameter-specific learning rates. The rates are adaptively adjusted via a ReduceLROnPlateau scheduler based on the compound loss function. 
Furthermore, the backpropagation step employs gradient clipping for stability.


\section{Experimental Results}
\label{sec:experiments}
\textbf{Currently, there are no learning-free or neural techniques for \textit{real-time} recalibration of \textit{infrared multi-camera} systems.} In the supplementary material, we adopt standard calibration and optimization techniques for infrared cameras and demonstrate the impracticality of using traditional methods for on-the-fly calibration. Below, we provide a comprehensive evaluation of our method's accuracy across various scenarios, demonstrating its suitability for time-sensitive, high-accuracy applications. We conclude with an assessment of the generalizability of our approach to diverse multi-camera system configurations and calibration objects.



In Table \ref{tab:point_based_results}, we detail the performance of our model under various conditions, reporting the average RMSE reprojection error RE$^{20K}_{avg}$ across three trials on synthetic test sets, each comprising 20,000 data samples. Our model demonstrates robust adaptability to different levels of perturbations, where performance gracefully degrades as perturbations reach extreme values. For example, with an O-shape camera system comprising 10 cameras ($N_{C}=10$ \Oshape{0.08}), and a calibration object with 8 fiducials ($N_{fid.} = 8$), our method achieves a RE$^{20K}_{avg}$ of $9.93\pm$ \SI{1.4e-4}{}, $11.53\pm$ \SI{1.5e-4}{}, and $14.21\pm$ \SI{1.8e-4}{}, for perturbations of up to $5\%, 10\%, 20\%$, respectively, as shown in rows 1-3. As we explain in the subsequent section, the range of motion is often constrained in real-world scenarios due to operational limitations. Nevertheless, our approach demonstrates robust performance and effectively predicts camera poses across a wide range of motion, i.e., $\theta \sim \mathcal{U}(0, 2\pi)$, $\phi \sim \mathcal{U}(0, \pi/2)$, and $\alpha \sim \mathcal{U}(0, 2\pi)$.

\begin{table*}[tb]
    \caption{\textbf{Experimental results.} \textbf{RE}$^{20K}_{avg}$ is the average RMSE reprojection error on 3 different trials (synthetic test sets, each comprising 20,000 data samples). Training on all models includes adding a \textbf{Max Perturb. $\kappa \in [\text{min}\%, \text{max}\%]$} to the OEM camera intrinsic and extrinsic parameters respectively. \textbf{$\boldsymbol{N_{fid.}}$} and \textbf{$\boldsymbol{N_C}$} are the number of 3D fiducials and the number of cameras, respectively. The rotation angle $\alpha$ remains the same in all experiments i.e., $\alpha \sim \mathcal{U}(0, 2\pi)$. The parameters for Pt and Img models are $\sim$33m and $\sim$86m, respectively. }
    \centering
    \resizebox{\textwidth}{!}{%
      \rowcolors{2}{white}{gray!25}
      \begin{tabular}{@{}c|c|c|c|c|c|c|c@{}}
        \toprule
        \textbf{\specialcell[t]{Row\\ \#}} &
         \textbf{Variant} &  \textbf{$\boldsymbol{\theta}$} &  \textbf{$\boldsymbol{\phi}$} & \specialcell[t]{\textbf{$\boldsymbol{N_{fid.}}$}\\(object)}  & \specialcell[t]{\textbf{$\boldsymbol{N_C}$}\\(config.)} & \specialcell[t]{\textbf{Max Perturb.}\\$\kappa_{int.}, \kappa_{ext.} \in [\text{min}\%, \text{max}\%]$} & \specialcell[t]{\textbf{RE}$^{20K}_{avg}$\\(pixels)}\\
        \midrule
        1 & Pt & $\sim \mathcal{U}(0, 2\pi) $ & $\sim \mathcal{U}(0, \pi/2)$ & 8 \cube{0.15} & 10 \Oshape{0.08} & $\pm2.5, \pm2.5$ & $9.93\pm$ \SI{1.4e-4}{}\\
        2 & Pt & $\sim \mathcal{U}(0, 2\pi)$ & $\sim \mathcal{U}(0, \pi/2)$ & 8 \cube{0.15} & 10 \Oshape{0.08} & $\pm5, \pm5$ & $11.53\pm$ \SI{1.5e-4}{}\\
        3 & Pt & $\sim \mathcal{U}(0, 2\pi)$ & $\sim \mathcal{U}(0, \pi/2)$ & 8 \cube{0.15} & 10 \Oshape{0.08} & $\pm10, \pm10$ & $14.21\pm$ \SI{1.8e-4}{}\\
        4 & Pt & $\sim \mathcal{U}(0, 2\pi)$ & $\sim \mathcal{U}(0, \pi/2)$ & 8 \cube{0.15} & 7 \Ushape{0.08} & $\pm2.5, \pm2.5$ & $14.67\pm$ \SI{3.5e-4}{}\\
        5 & Pt & $\sim \mathcal{U}(0, 2\pi)$ & $\sim \mathcal{U}(0, \pi/2)$ & 8 \cube{0.15} & 4 \Tshape{0.08} & $\pm2.5, \pm2.5$ & $12.89\pm$ \SI{2.1e-4}{}\\
        6 & Pt & $\sim \mathcal{U}(0, 2\pi)$ & $\sim \mathcal{U}(0, \pi/2)$ & 64 \sphere{0.15} & 10 \Oshape{0.08} & $\pm2.5, \pm2.5$ & $14.92\pm$ \SI{7.3e-5}{}\\
        7 & Pt & 0 & 0 & 8 \cube{0.15} & 6 \OshapeSix{0.08} & $0,0$ & $1.12\pm$ \SI{5e-7}{}\\
        8 & Pt & 0 & 0 & 8 \cube{0.15} & 6 \OshapeSix{0.08} & $\pm2.5, 0$ & $3.08\pm$ \SI{9.5e-6}{}\\
        9 & Pt & 0 & 0 & 8 \cube{0.15} & 6 \OshapeSix{0.08} & $\pm10, 0$ & $8.04\pm$ \SI{7.2e-4}{}\\
        10 & Img & 0 & 0 & 8 \cube{0.15} & 6 \OshapeSix{0.08} & $\pm2.5, 0$ & $4.55 \pm$ \SI{1.8e-4}{}\\
        11 & Img & 0 & 0 & 8 \cube{0.15} & 10 \Oshape{0.08} & $\pm2.5, \pm2.5$ & $4.12 \pm$ \SI{2.4e-6}{}\\
        12 & Img & 0 & 0 & 8 \cube{0.15} & 10 \Oshape{0.08} & $\pm5, \pm5$ & $6.01 \pm$ \SI{1.6e-5}{}\\
        13 & Img & 0 & 0 & 8 \cube{0.15} & 10 \Oshape{0.08} & $\pm10, \pm10$ & $6.97 \pm$ \SI{1.1e-5}{}\\
      \bottomrule
      \end{tabular}
      \label{tab:point_based_results}
      }
\end{table*}

\subsection{Generalization to Arbitrary Configurations \& Calibration Objects} We conducted comprehensive experiments with various camera configurations and calibration objects to evaluate the generalization of our method. Cameras were positioned in geometric configurations common to real-world scenarios, such as the O-shaped ($N_{C}=10$ \Oshape{0.08}), U-shaped ($N_{C}=7$ \Ushape{0.08}) and T-shaped ($N_{C}=4$ \Tshape{0.08}) arrangements. As shown in Table \ref{tab:point_based_results} (in rows 1, 4, and 5), even with a small number of fiducials i.e., $N_{fid.} = 8$, the reprojection errors RE$^{20K}_{avg}$ are low with $9.93\pm$ \SI{1.4e-4}{}, $14.67\pm$ \SI{3.5e-4}{}, $12.89\pm$ \SI{2.1e-4}{}, for the O-shape, U-shape, and T-shape configurations, respectively. 

The versatility of our approach is further demonstrated through tests involving calibration objects with different shapes and numbers of fiducials. Specifically, we evaluated our model using a calibration cube with $N_{fid.}=8$ and a calibration sphere with $N_{fid.}=64$ fiducials. As shown in Table \ref{tab:point_based_results}(rows 1, 6), our method maintains a consistent error profile for both calibration objects. 

\subsection{Comparison with CMM-calibrated multi-camera system.} As previously stated, it is important to recognize that regressing camera poses under the conditions that $\theta \sim \mathcal{U}(0, 2\pi)$, $\phi \sim \mathcal{U}(0, \pi/2)$ is more challenging than in typical real-world scenarios where multi-camera systems have predefined operational ranges for $\theta$ and $\phi$. Our method, when practically tested in a surgical setting using a multi-camera system with a predefined operational range of motion (specifically, an overhead fixed O-shaped system ($N_{C}=6$ \OshapeSix{0.08}) where $\theta=\phi=0$, and a calibration object with $N_{fid.}=8$), demonstrated superior performance. This system was previously calibrated (offline) 
using a high-end 
CMM. Our approach surpassed the CMM calibration in terms of reprojection error within this motion range.
CMM calibration parameters \textit{without perturbations} led to an average reprojection error of 1.80 pixels. 
In contrast, our method led to an average reprojection error of $1.12\pm$ \SI{5e-7}{} on a model trained \textit{without perturbations} (Table \ref{tab:point_based_results}; row 7), and an average reprojection error of $3.08\pm$\SI{9.5e-6}{} (Table \ref{tab:point_based_results}; row 8) and $8.04\pm$ \SI{7.2e-4}{} (Table \ref{tab:point_based_results}; row 9) for \textit{perturbations of up to 5\% and 20\%}, respectively.  
Additionally, with an average inference time of 0.0026 seconds on a system equipped with a Nvidia RTX 4090 GPU, our method proves capable of supporting real-time applications.

\section{Conclusion}
Our work presents a neural calibration method tailored for the real-time and adaptive calibration of multi-camera systems. Central to our approach is the combination of dynamic camera pose synthesis with a differentiable projection model, which facilitates the direct optimization of camera parameters from image data. 
Comprehensive experimental analysis demonstrated the robustness of the method and its capacity to accurately predict calibration parameters while accommodating random perturbations. 

We further elaborated on the practicality of our method, contrasting the impracticality of recalibrating or optimizing camera parameters at each step in real-time applications.
 Our evaluations revealed our method's superior accuracy and generalizability across different scenarios. 
The introduction of two variants—catering to systems with either direct 2D projections of 3D fiducials or color-coded 2D projected points—proves our method's flexibility and broad applicability in diverse operational contexts. Future research directions include the integration of perturbations into the extrinsic parameters of the camera models. This holds the potential to significantly broaden the applicability of our approach in more dynamically varied environments.

{\small
\bibliographystyle{ieee_fullname}
\bibliography{egbib}
}

\section*{Supplementary Material}

In the supplementary material, we present details on the network architectures and an ablation on their components, comparisons with standard calibration and optimization techniques, qualitative results and training nuances of the real-time neural multi-camera system calibration method. Additionally, through a series of visualizations, we illustrate the precision of our model in predicting camera poses against ground truth, with an emphasis on robustness in the presence of perturbations. Lastly, we address the limitations and discussions. 

\section{Network Architecture}
We introduce two variants: the first one, known as the point-based model (Pt), is intended for multi-camera systems with onboard processing that directly outputs the 2D projections of the 3D fiducials, which is the typical case for IR multi-camera systems; the second one is the image-based model (Img), designed for multi-camera systems that output images for subsequent processing, rather than the point coordinates. The architecture of both variants is summarized in the supplementary material.

\subsection{Point-based variant.}
Given an input tensor $X \in \mathbb{R}^{N_{C} \times N_{fid.} \times 2}$, where $N_{C}$ is the number of cameras, and $N_{fid.}$ is the number of fiducials, the point-based model performs the following operations:
\begin{enumerate}
    \item \textbf{Embedding Layer $f_{embed}$:} Maps input fiducial points to a higher-dimensional space: $X_{enc} = f_{embed}(X)$, where $X_{enc} \in \mathbb{R}^{N_{C} \times 512}$.
    \item \textbf{Camera Identity Encoding $CIE$:} Incorporates camera-specific information: $X_{enc}^{'} = X_{enc} + CIE(N_{C}, 512)$, where $X_{enc}^{'} \in \mathbb{R}^{N_{C} \times 512}$.
    \item \textbf{Transformer Encoder $f_{TE}$:} Processes embeddings through a transformer encoder to model complex relationships: $X_{enc}^{''} = f_{TE}(X_{enc}^{'})$, $X_{enc}^{''} \in \mathbb{R}^{N_{C} \times 512}$.
    \item \textbf{Downstream Heads $h_{*}^{M}$:} Process the encoder output through separate prediction heads for different outputs $X_{out}^{*} \in \mathbb{R}^{M}$ (rotation $r_{6D}$, translation $t$, focal lengths $fc = (f_{x}, f_{y})$, principal point $pp = (c_{x}, c_{y})$, and distortion coefficients $kc = (kc_{1,2,3}, p_{1,2})$):
    \begin{align*}
    X_{out}^{r_{6D}} = h_{R}^{6}(X_{enc}^{''}),& \quad X_{out}^{t} = h_{t}^{3}(X_{enc}^{''}), \\
    X_{out}^{fc} = h_{fc}^{2}(X_{enc}^{''}), \quad X_{out}^{pp} = &h_{pp}^{2}(X_{enc}^{''}), \quad X_{out}^{kc} = h_{kc}^{5}(X_{enc}^{''})
    \end{align*}
    \item In a final step, the outputs $X_{out}^{*}$ are combined to form the output tensor $\hat{X}_{out}$ representing the combined camera parameters: \[ \hat{X}_{out} = (\gamma(X_{out}^{r_{6D}}), X_{out}^{t}, X_{out}^{fc}, X_{out}^{pp}, X_{out}^{kc})\] where $\hat{X}_{out} \in \mathbb{R}^{N_{C} \times 21}$, $\gamma(.)$ is a function that expands the 6D parameterized rotation to a rotation matrix as described in Section \textcolor{red}{3} in the main paper, and 21 is the number of calibration parameters per camera.
\end{enumerate}


\noindent Incorporating camera identity encoding $CIE$ into our architecture significantly improves performance by disambiguating camera perspectives, as shown in the ablations in Section \ref{subsec:ablation}. The $CIE$ applies a simple one-hot encoding strategy where each camera is assigned a unique identifier in this embedding space. To accommodate scenarios where the number of cameras exceeds the embedding dimension, it introduces a small amount of noise to each encoding, ensuring that each camera's encoding remains distinct. This enhances robustness and accuracy for spatially-aware tasks and supports generalization to new configurations and arbitrary camera setups. 


\subsection{Image-based variant.}
The image-based variant utilizes a Vision Transformer (ViT) to process the input image tensor $X_{in} \in \mathbb{R}^{N_{C} \times Ch \times H \times W}$ and predict camera parameters. 

\begin{enumerate} 
    \item \textbf{Encoder $f_{encoder}$.} The input image tensor is resized to $224 \times 224$ pixels and passed through a ViT encoder. The ViT model is adapted by replacing its classification head with an identity layer, allowing the model to output a feature representation directly $X_{enc} = f_{encoder}(X_{in})$, where $X_{in} \in \mathbb{R}^{N \times 3 \times 224 \times 224}$, and $X_{enc} \in \mathbb{R}^{N_{C} \times 768}$.
    
    \item \textbf{Bottleneck Layer $f_{bottleneck}$.} The encoded features are further processed through a bottleneck layer to reduce dimensionality and focus on relevant features for parameter prediction:
    $ X_{bottleneck} = f_{bottleneck}(X_{enc})$, and comprises a linear layer, reducing features to $\mathbb{R}^{N_{C} \times 128}$.
    
    \item \textbf{Downstream Heads:} Similar to the point-based model variant, separate heads are used for predicting the camera parameters
    from the bottleneck features $X_{bottleneck}$, and combined to form the output tensor $\hat{X}_{out} \in \mathbb{R}^{N_{C} \times 21}$.
\end{enumerate}

\begin{figure*}[!ht]
    \centering
    \includegraphics[width=0.95\textwidth]{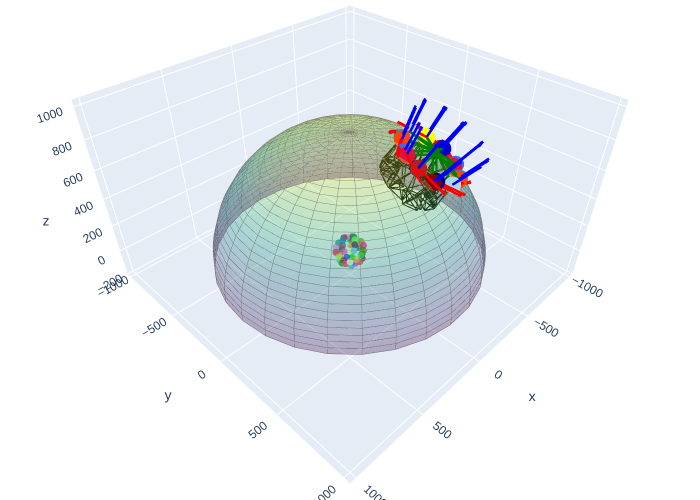}
    \caption{\textbf{Visualization of predicted vs ground truth camera poses.} The calibration object is a sphere with 64 fiducials. The multi-camera system configuration is O-shaped comprising 10 cameras. For closer inspection please refer to the interactive visualization in the \textbf{cameras.html} file.}
    \label{fig:qualitative_result}
\end{figure*}

\subsection{Ablation Study}
\label{subsec:ablation}

We conducted an ablation study to demonstrate the impact and significance of individual components of our network. All models have been trained following the training strategy in Section \textcolor{red}{4.4} in the main paper, given the conditions explained in Table \ref{tab:ablation_results}.

\noindent
\subsubsection{Impact of Value Scaling} We highlight the significant role of scaling rotation and distortion coefficients prior to RMSE loss calculation. Given the inherently small magnitudes of these coefficients—elements of the rotation matrix ($0 < r_{i} < 1$) and distortion parameters ($kc_{1,2,3}, p_{1,2} \lesssim 0.1$)—their contribution to the overall loss is minimal. Introducing a scaling factor, $\lambda_{scale}$, amplified their effect, as evidenced by the increased prediction errors observed when omitting this factor, detailed in Table \ref{tab:ablation_results}.

\noindent
\subsubsection{Camera Identity Encoding $CIE$} We showcase the significance of Camera Identity Encoding ($CIE$). The results in Table \ref{tab:ablation_results} clearly indicate that incorporating $CIE$ leads to notable performance improvements. By integrating CIE, our model distinctly differentiates between camera perspectives, leading to marked improvements in accuracy 
for spatially-aware tasks.

\noindent
\subsubsection{Impact of Encoder} We contrast the performance impact between utilizing a transformer encoder and a 1D convolutional (CNN-1D) encoder within our model. As detailed in Table \ref{tab:ablation_results}, 
the integration of a transformer encoder significantly enhances model accuracy. Similarly, for the image-based model, the ViT encoder results in increased precision when compared with CNN-2D. 

\begin{table*}[tb]
  \caption{ \textbf{Ablation study results}. Training on all models includes a perturbation of \textbf{Max Perturb. $\kappa \in [\text{-2.5}\%, \text{+2.5}\%]$} of the OEM intrinsic parameters, utilizing 10 cameras ($N_{C}=10$) and a calibration object with 8 fiducials ($N_{fid.} = 8$). Pt models have been trained for 300k epochs and Img models for 150k epochs.}
  \centering
  \setlength{\tabcolsep}{8pt}
  \resizebox{\textwidth}{!}{%
      \rowcolors{2}{white}{gray!25}
      \begin{tabular}{c|c|c|c|c|c|c|c|c}
        \toprule
        \textbf{\specialcell[t]{Row\\ \#}} &  \textbf{Variant}& \specialcell[t]{\#\\ \textbf{Params}} & \textbf{$\boldsymbol{\theta}$} &  \textbf{$\boldsymbol{\phi}$} & \textbf{Encoder} & \specialcell[t]{\textbf{Scaled}\\$\lambda_{scale}=1000$} & \specialcell[t]{\textbf{Camera}\\\textbf{Identity}\\ \textbf{Encoding}\\ $CIE$} &  \specialcell[t]{\textbf{RE}$^{20K}_{avg}$\\(pixels)}\\
        \midrule
        1 & Pt & $\sim$ 33m & $\sim \mathcal{U}(0, 2\pi)$ & $\sim \mathcal{U}(0, \pi/2)$ & Transformer & \checkmark & \checkmark & $12.47 \pm$ \SI{6e-4}{} \\
        2 & Pt & $\sim$ 33m & $\sim \mathcal{U}(0, 2\pi)$ & $\sim \mathcal{U}(0, \pi/2)$ & Transformer & & \checkmark & $421.26 \pm$ \SI{9.6e-3}{} \\
        3 & Pt & $\sim$ 33m & $\sim \mathcal{U}(0, 2\pi)$ & $\sim \mathcal{U}(0, \pi/2)$ & Transformer & \checkmark &  & $25.49 \pm$ \SI{2e-2}{}  \\
        4 & Pt & $\sim$ 8m & $\sim \mathcal{U}(0, 2\pi)$ & $\sim \mathcal{U}(0, \pi/2)$ & CNN-1D & \checkmark & \checkmark & $96.93 \pm$ \SI{3.9e-2}{} \\
        5 & Img & $\sim$ 86m & 0 & 0 & ViT & \checkmark & - & $6.89 \pm$ \SI{5.8e-5}{} \\
        6 & Img & $\sim$ 24m & 0 & 0 & CNN-2D & \checkmark & - & $20.62\pm$ \SI{2.8e-5}{} \\
        
      \bottomrule
      \end{tabular}
      \label{tab:ablation_results}
  }
\end{table*}

\section{Comparison with Standard Calibration and Optimization Techniques}
\label{sec:supp_experiments}
In this section, we explore the differences between recalibrating and optimizing infrared cameras at every step in real-time applications. We start our discussion in Section \ref{subsec:run_time}, examining the drawbacks of recalibrating at each step using standard methods. Through experiments, we demonstrate the significant time required for this process, thereby illustrating the impracticality of using traditional methods for on-the-fly calibration in multi-camera systems. Following this, Section \ref{subsec:no_action} presents experiments that emphasize the difficulties in minimizing reprojection errors using Bundle Adjustment (BA). 

\subsection{Inference vs. Recalibration}
\label{subsec:run_time}

\begin{figure}[!ht]
    \centering 
    \includegraphics[width=0.45\textwidth]{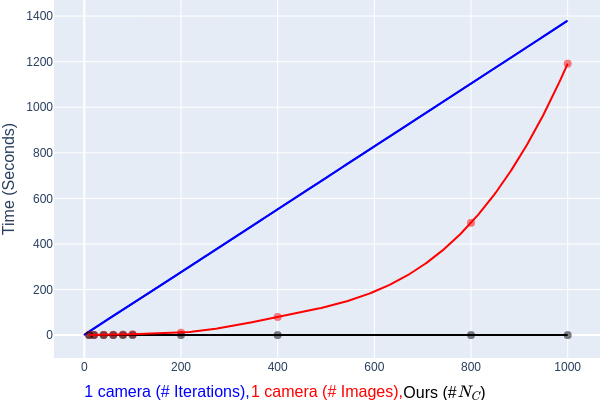}
    \caption{\textbf{Runtime for traditional camera calibration.} Exponential growth in calibration time with increasing number of images (red; 1 camera). Linear increase w.r.t. LM iterations on 100 images (blue; 1 camera). Ours; real-time ($\tau_{min}^{i=1}=0.0026s$, $\tau_{max}^{i=10}=0.012s$) for increasing number of cameras $2^{i}, 1 \leq i \leq 10$ (black).}
    \label{fig:opencv_runtime}
\end{figure}

First, we address the limitations of traditional calibration methods. 
Traditional calibration methods require a large number of images, especially for applications demanding high accuracy. In such scenarios, the requirement can escalate to thousands of images, leading to a prohibitive increase in execution time (i.e., $> 1000$ images for vision-based computer-assisted surgical multi-camera systems). Even in the hypothetical scenario where capturing thousands of images instantaneously were possible, the calibration process itself remains time-consuming and scales poorly as image quantity increases. As depicted in Figure \ref{fig:opencv_runtime} for a single camera, this execution time grows exponentially with the number of images, rendering a recalibration at every step impractical within real-time applications. Additionally, increasing the number of iterations in the Levenberg-Marquardt refinement step leads to a linear time increase, as illustrated by the blue curve in Figure \ref{fig:opencv_runtime}.

In contrast, our method operates on a fixed number of cameras, $N_{C}$, which inherently caps the inference time to $\tau_{min}^{i=1}=0.0026s$, $\tau_{max}^{i=10}=0.012s$ for increasing number of cameras $2^{i}, 1 \leq i \leq 10$ on an Nvidia RTX 4090, as shown in Figure \ref{fig:opencv_runtime} by the black curve. This translates to $\sim 387$ inferences per second, demonstrating a significant advantage in the context of real-time applications. 

\subsection{Inference vs. Minimizing Reprojection Errors}
\label{subsec:no_action}

We now pivot our experiments to investigate the feasibility of minimizing reprojection error at every step using Bundle Adjustment (BA) as an alternative to recalibrating at every step, in the context of real-time optimization
amidst decalibration.

We conducted 1000 trials, each applying random perturbations within a range of $\kappa \in [-10\%, 10\%]$ to the OEM intrinsic parameters. For every trial, our objective was to minimize the reprojection error with BA, utilizing data from $N_{fid.} = 8$ points and $N_{C} = 6$ cameras. 

Our findings highlight a significant constraint: the time taken for a single BA iteration, averaging 0.0385 seconds, substantially exceeds our execution time of 0.0026 seconds for a single inference, and proves that optimizing camera parameters by minimizing reprojection errors at every step is not a viable choice for real-time applications. 

\begin{figure}[!ht]
    \centering 
    \includegraphics[width=0.45\textwidth]{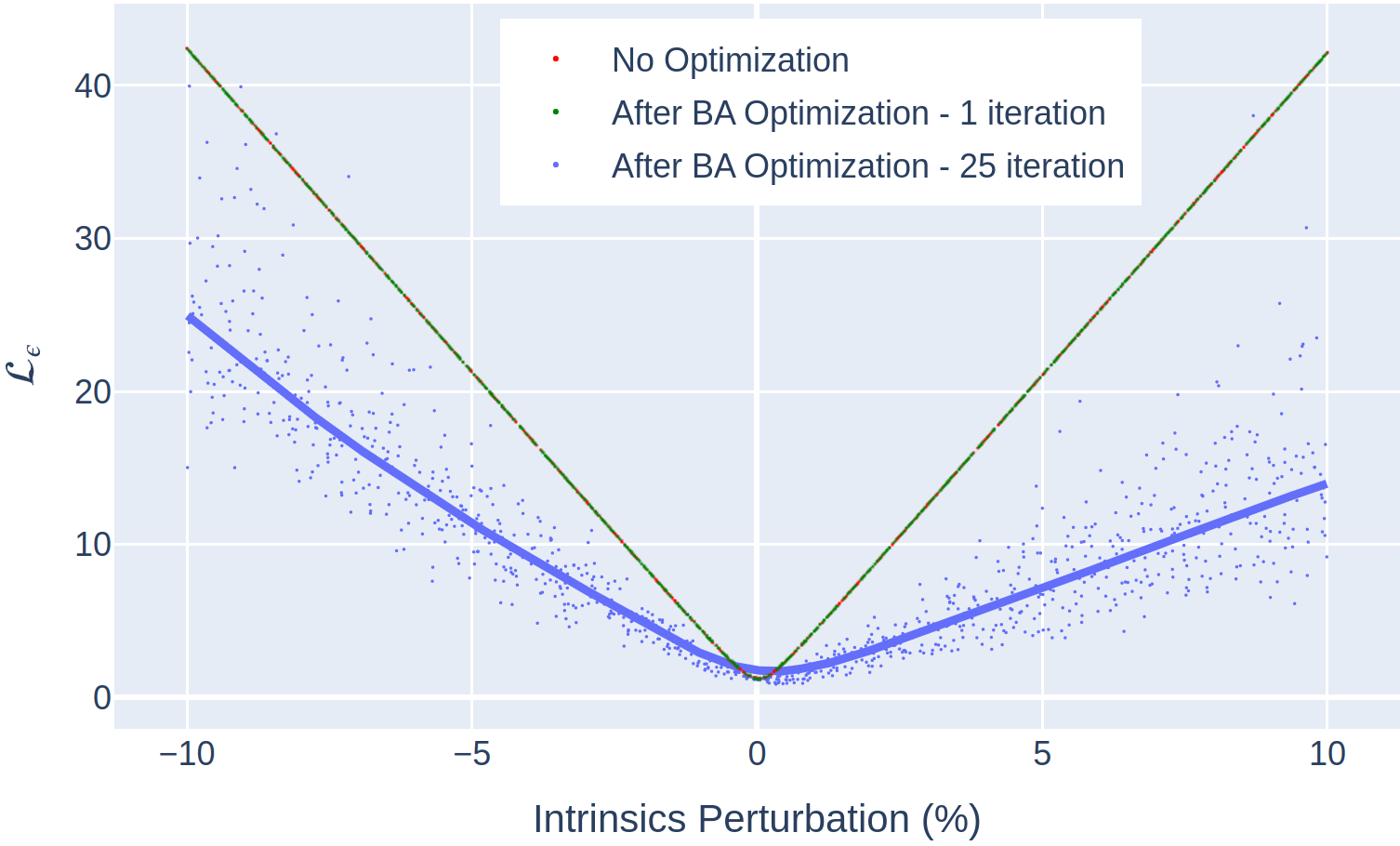}
    \caption{\textbf{Effect of decalibration on reprojection error (RMSE).} OEM intrinsic parameters are perturbed by 20\%, i.e., $\kappa \in [-10\%, 10\%]$, simulating potential decalibration. Reprojection error without intervention (red), with 1 iteration of Bundle Adjustment (green), with 25 iterations (blue); 1000 trials. }
    \label{fig:decalibraton_vs_reprojection}
\end{figure}

Furthermore, a single iteration of BA resulted in minimal improvement in reprojection error, as illustrated in Figure \ref{fig:decalibraton_vs_reprojection} in green vs red. Subsequent experiments showed that achieving a noticeable reduction in reprojection error required an average of 25 BA iterations per system's capture, culminating in an untenable average execution time of 1.2892 seconds for real-time processing. 
Despite these efforts, the reprojection error remained significantly high (Figure \ref{fig:decalibraton_vs_reprojection}; blue), compared to our method (Table \textcolor{red}{1} in the main paper; rows 9, 10).

\section{Qualitative Results}
In Figure \ref{fig:qualitative_result}, we show a qualitative result of the calibration from the multi-camera system calibration. The blue-shaded spheres represent the ground truth values. The red-shaded spheres represent the predicted values. The single yellow sphere per multi-camera system represents the first camera, and is singled out for visualization purposes only, to demonstrate the in place rotation $R_{random}$ i.e. that the multi-camera system does not always have the same vertical orientation. For closer inspection please refer to the interactive visualization in the supplementary material. 

Figure \ref{fig:reprojected_fiducials} illustrates the reprojection accuracy of 3D fiducials using predicted camera poses (white dots)  contrasted against the ground-truth projections (color-coded). In this example, a point-based model was trained with 5\% perturbation, and the input points to the model were generated with \textit{no perturbation} to the camera parameters. The multi-camera system is arranged in an O-shape, consisting of 10 cameras, with a cube with 8 fiducials as the calibration object.

\begin{figure*}[!ht]
\centering

\includegraphics[width=\textwidth]{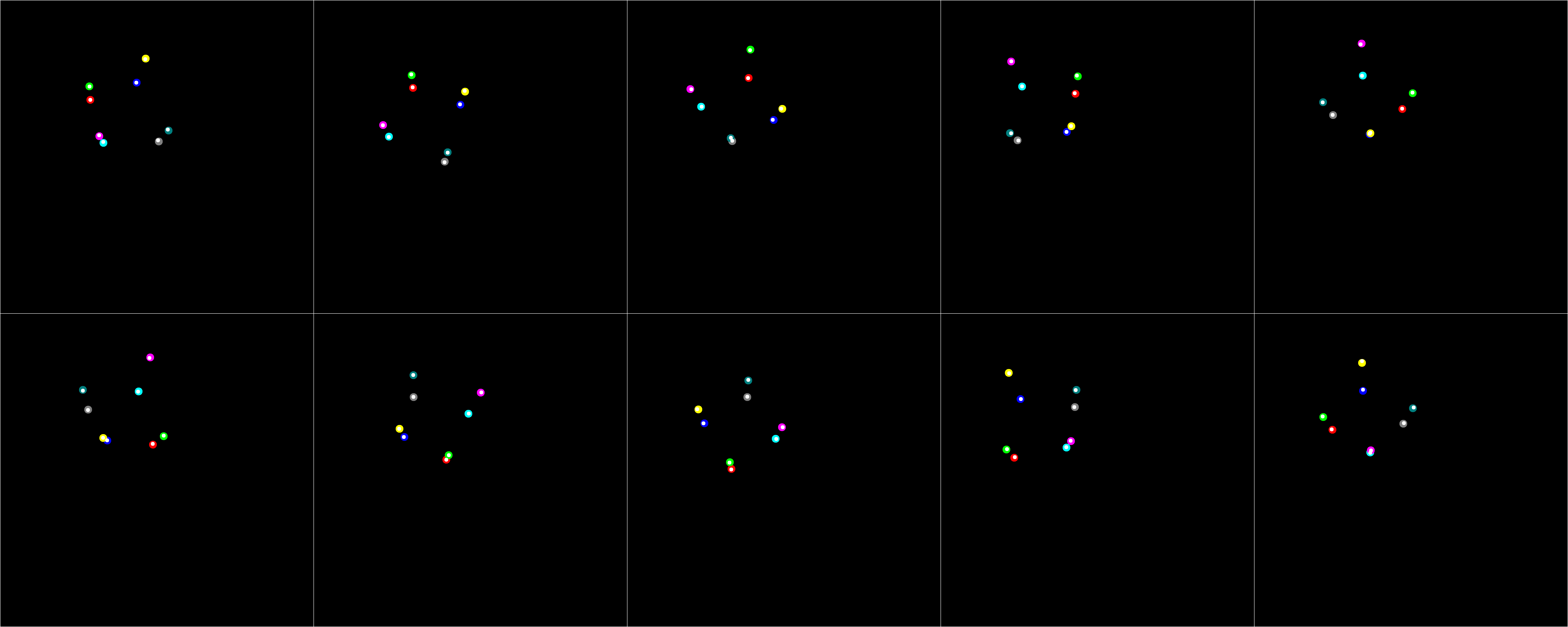}
\caption{\textbf{Reprojection of 3D Fiducials with Predicted Camera Poses (With 5\% Perturbation).} This figure illustrates the resilience and accuracy of our pose estimation model during decalibration of the camera parameters of up to 5\% perturbation. The calibration cube and the O-shaped arrangement of 10 cameras remain constant as in Figure \ref{fig:reprojected_fiducials}, allowing for a direct comparison across different testing conditions. Note: The reprojected points are shown in white. For optimal visibility, please zoom in.}
\label{fig:reprojected_fiducials_w_pert}
\end{figure*}

\begin{figure*}[!hb]
\centering
\includegraphics[width=\textwidth]{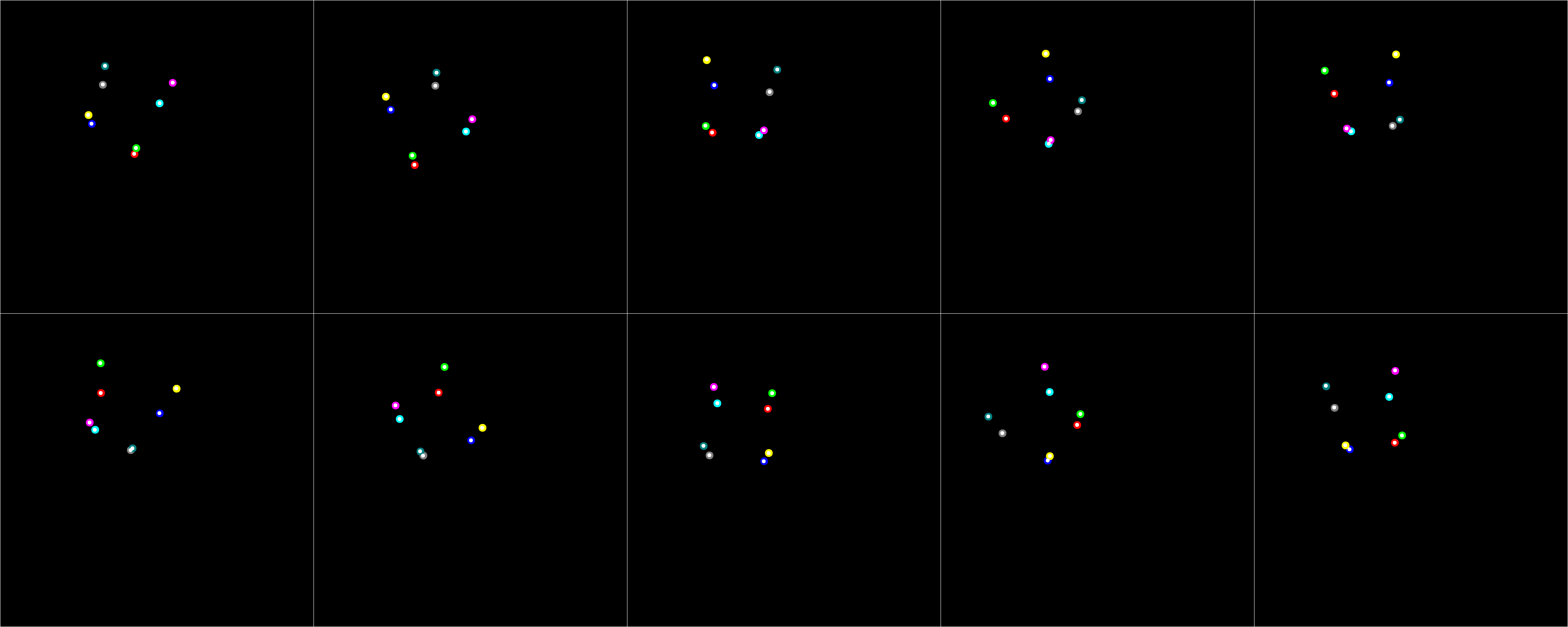}
\caption{\textbf{Reprojection of 3D Fiducials with Predicted Camera Poses (Test Data Generated Without Perturbation).} This figure demonstrates the reprojection accuracy in a O-shaped multi-camera setup comprising 10 cameras, with the calibration object being a cube with 8 fiducials. The comparison between the predicted (in white) and ground-truth (color-coded; enlarged for visualization purposes) projections demonstrates the precision of our model in the absence of perturbation. Note: The reprojected points are shown in white. For optimal visibility, please zoom in.}
\label{fig:reprojected_fiducials}
\end{figure*}

Figure \ref{fig:reprojected_fiducials_w_pert} shows the same model as Figure \ref{fig:reprojected_fiducials}, i.e., trained with a 5\% perturbation. However, in this example we simulate a decalibration by \textit{generating the test data from perturbed camera parameters of up to 5\%}. Beyond the quantitative evaluation presented in the previous section, these figures facilitate a direct qualitative comparison of model performance and clearly demonstrate the robustness of our pose estimation technique under conditions of uncertainty.

\section{Training: Less Effective Strategies}
In this section, we describe our exploration of various loss and regularization terms intended to improve model performance. Despite their theoretical potential, these methods did not enhance our results in practice.

\subsection{Regularization terms}
\begin{itemize}
    \item A regularization term, denoted as $\mathcal{L}_{\kappa_{c}}$, was introduced to encourage minimal distortion coefficients ($k_1$, $k_2$, $k_3$, $k_4$, $k_5$) by minimizing the sum of the absolute values of these coefficients.
    \item A regularization term, $\mathcal{L}_{f{c}}$, was designed to ensure consistency in the focal lengths $f_x$ and $f_y$ by minimizing the squared difference between these focal lengths.
    \item The regularization term $\mathcal{L}_{pp}$ aimed to align the principal points ($c_x$, $c_y$) with the image center by minimizing the squared Euclidean distance between the principal points and the center of the image.
    \item A regularization term, $\mathcal{L}_{largefc}$, is formulated to discourage small focal lengths by imposing a penalty on the inverse of the sum of the focal lengths along both the x and y axes.
    \item Applying an L1 norm regularization to our model's parameters aimed to encourage sparsity and reduce overfitting by penalizing large weights, thus simplifying the model. This technique, intended to improve generalization, unfortunately did not yield the expected performance enhancements in our camera parameter prediction task. 
\end{itemize}

\subsection{Losses}
We investigated the log-cosh loss function, attracted by its theoretical benefits such as smooth gradients, outlier robustness, and balanced error sensitivity. These characteristics suggested that log-cosh could enhance prediction precision and stability. The log-cosh loss is given by,
\begin{align*}
 \mathcal{L}_{logcosh} = \frac{1}{N} \sum_{i=1}^{N} &( (y_i - \hat{y}_i) \tanh(y_i - \hat{y}_i) -  \\ 
                &\log(2) + \log\left(1 + e^{-2 |y_i - \hat{y}_i|}\right) ) 
\end{align*}
where $y_i$ represents the target values, and $\hat{y}_i$ represents the predicted values. However, contrary to our expectations, empirical testing revealed that log-cosh loss did not outperform our existing loss function described in Section \textcolor{red}{4.3} in the main paper.

\begin{figure*}[!ht]
    \centering
    \includegraphics[width=\textwidth]{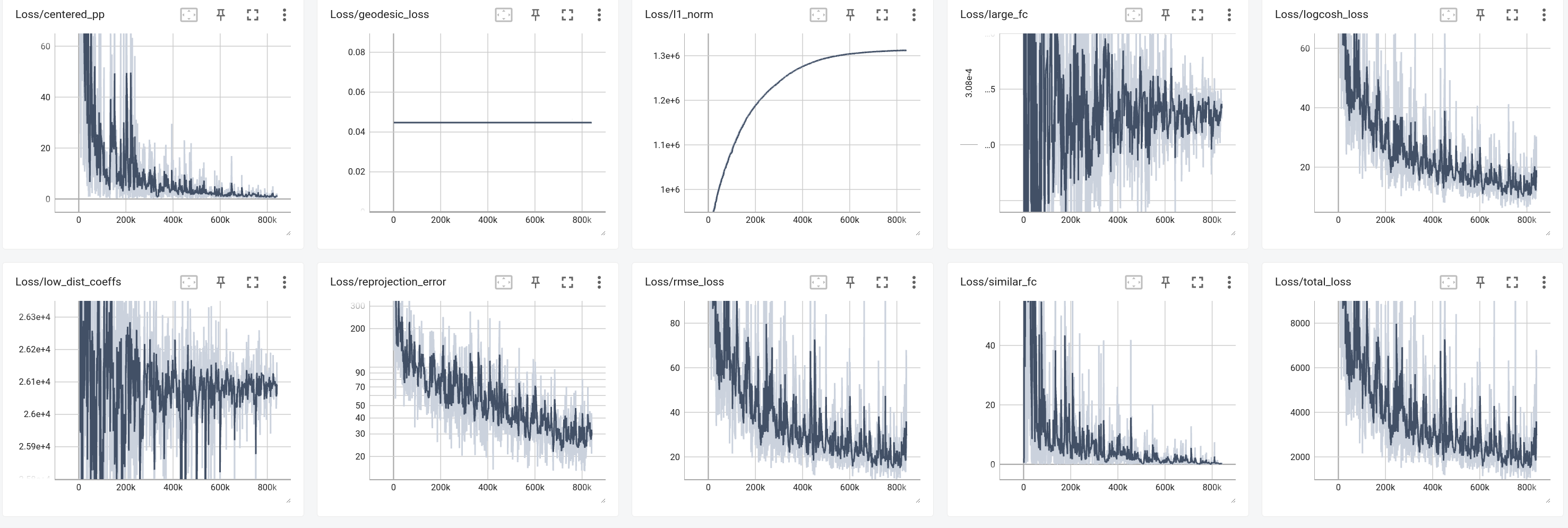}
    \caption{\textbf{Illustration of the proposed loss compared to auxiliary regularization and loss terms.} This figure demonstrates that minimizing the proposed loss leads to a reduction in auxiliary losses. However, this is a non-reciprocal relationship where the inclusion of auxiliary losses does not further improve model performance beyond the capabilities of the proposed loss which is a combination of the reprojection error, the RMSE error of the parameters, and the geodesic error. Here, we present the training loss graphs for the point-based variant. As previously described, the distortion coefficients are scaled by scaling factor $\lambda_{scale}=1000$.}
    \label{fig:losses}
\end{figure*}

Figure \ref{fig:losses} illustrates a key observation: minimizing our proposed final loss also reduces the auxiliary losses we initially investigated-except from the L1 norm which relates to the network weights rather than the calibration error, and whose purpose is to enforce sparsity. However, this effect is not reciprocal; directly incorporating these auxiliary terms into our model did not further enhance performance beyond the improvements achieved with the final loss alone described earlier. This indicates that while our final loss effectively captures the essence of the auxiliary losses, adding them explicitly does not provide additional benefits. 
In the figure, we present the training loss patterns for the point-based variant. Comparable trends are observed with the image-based variant. 

\section{Limitations \& Discussion}
\subsection{Adhering to Operational Specifications}
Our camera pose synthesis methodology is specifically designed to meet the manufacturer's usage requirements, prioritizing high-accuracy predictions in real-world applications, particularly in time-critical contexts such as surgical applications. An important component of our methodology involves simulating a field of view that ensures comprehensive coverage of the area of interest, closely mimicking the actual setup. This is accomplished by ensuring a constant radius for the hemisphere, in accordance with the operational guidelines of the multi-camera system. Unlike some multi-camera systems that are statically affixed with no or limited range of motion, our methodology introduces a significant enhancement by allowing an extended range of motion—albeit at a fixed distance—over the entire hemisphere. The decision to adhere to the specified fixed radius is a deliberate design choice aimed at ensuring that our synthesized camera poses accurately reflect the real-world configuration.

\subsection{Limited Range of Motion for Image-based} Our assessment of the image-based variant shows promising results 
in multi-camera setups having a limited range of motion. However, to achieve broader generalization across diverse vantage points -especially at glazing angles-, a more advanced architecture is imperative. This is based on the observation that accuracy drops when using the full range of motion on the hemisphere, attributed to the complexity of input images. Specifically, the challenge arises as only a fraction, $N_{fid.}$ pixels, are relevant to the task amidst the backdrop of the remaining $H\times W-N_{fid.}$ pixels, which significantly narrows the dataset's utility for network learning.

\subsection{Model Customization} Our methodology demonstrates a tailored approach, achieving high accuracy in predicting camera parameters when applied to the particular multi-camera system, calibration object, and operating distance utilized during its training phase. This high level of accuracy reflects how well the method is customized to fit its specific training conditions, including the perturbations, up to the extent the model has been trained to withstand. However, it is important to note that the model's performance is finely tuned to this particular setup. Any deviation from the original camera setup, use of a different calibration object, or change in the desired maximum perturbation level requires the training of a new model tailored to those new conditions. 

\subsection{Decalibration Detection} Our method also extends beyond real-time calibration to effectively identify decalibration. By continuously comparing the predicted calibration parameters in real-time with those of the OEM, one can swiftly identify any deviations indicative of decalibration. This dual functionality positions our method as both a calibration tool and an operational integrity monitor, ensuring continuous accuracy and reliability of multi-camera systems through early detection and prompt recalibration response.

\end{document}